\documentclass[10pt,twocolumn,letterpaper]{article}

\usepackage{wacv}
\usepackage{times}
\usepackage{epsfig}
\usepackage{graphicx}
\usepackage{amsmath}
\usepackage{amssymb}
\usepackage{xcolor}
\usepackage{lipsum}  
\usepackage{caption}
\usepackage{algorithm}
\usepackage{algpseudocode}
\usepackage{verbatim}
\usepackage{booktabs}
\usepackage{subcaption}
\usepackage{array}
\usepackage{duckuments}
\usepackage{xcolor,colortbl}
\usepackage{multirow}
\usepackage{float}
\usepackage{comment}
\usepackage{soul}
\usepackage{appendix} 

\definecolor{onlineBlue}{rgb}{0.0, 0.48, 0.65}
\definecolor{onlineRed}{RGB}{255, 177, 177}
\definecolor{cerulean}{rgb}{0.0, 0.48, 0.65}
\definecolor{azure}{RGB}{0.0, 0.5, 1.0}

\newcommand{\std}[1]{{\footnotesize$\pm$#1}}

\def\wacvPaperID{675} 

\wacvfinalcopy 

\ifwacvfinal
\def\assignedStartPage{1} 
\fi

\usepackage[pagebackref=true,breaklinks=true,colorlinks,bookmarks=false]{hyperref}

\ifwacvfinal
\else
\fi

\begin{document}

\title{Self-Supervised Generative Style Transfer for One-Shot Medical Image Segmentation}

\author{Devavrat Tomar$^{1}$ \quad 

Behzad Bozorgtabar$^{1,2}$ \quad \\
 
Manana Lortkipanidze \quad  

Guillaume Vray$^{1}$ \quad 

Mohammad Saeed Rad$^{1}$ \quad 

Jean-Philippe Thiran$^{1,2}$
\and
$^1$LTS5, EPFL, Switzerland \quad \quad
$^2$CIBM, Switzerland \\
{\tt\small \{devavrat.tomar, firstname.lastname\}@epfl.ch} \quad \quad
}




\twocolumn[{%
\renewcommand\twocolumn[1][]{#1}%
\maketitle
\begin{center}
    \centering
    \vspace{-2mm}
    \includegraphics[width=\textwidth]{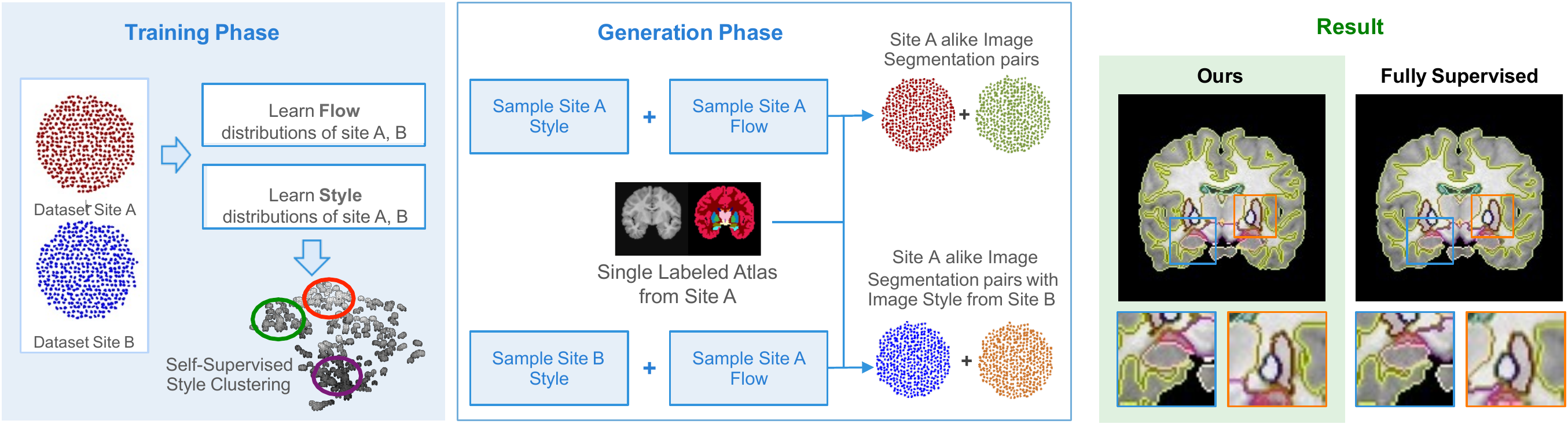}
    \captionof{figure}{\textbf{Cross-site adaptation of the proposed one-shot segmentation vs. Fully Supervised segmentation}. Our method uses volumetric self-supervised learning for style transfer by leveraging unlabeled data. Zoom in for the best view.}\label{fig:eye_catching}
\end{center}%
}]

\maketitle


\begin{abstract}\label{abstract}
In medical image segmentation, supervised deep networks' success comes at the cost of requiring abundant labeled data. While asking domain experts to annotate only one or a few of the cohort's images is feasible, annotating all available images is impractical. This issue is further exacerbated when pre-trained deep networks are exposed to a new image dataset from an unfamiliar distribution. Using available open-source data for ad-hoc transfer learning or hand-tuned techniques for data augmentation only provides suboptimal solutions. Motivated by atlas-based segmentation, we propose a novel volumetric self-supervised learning for data augmentation capable of synthesizing volumetric image-segmentation pairs via learning transformations from a single labeled atlas to the unlabeled data. Our work's central tenet benefits from a combined view of one-shot generative learning and the proposed self-supervised training strategy that cluster unlabeled volumetric images with similar styles together. Unlike previous methods, our method does not require input volumes at inference time to synthesize new images. Instead, it can generate diversified volumetric image-segmentation pairs from a prior distribution given a single or multi-site dataset. Augmented data generated by our method used to train the segmentation network provide significant improvements over state-of-the-art deep one-shot learning methods on the task of brain MRI segmentation. Ablation studies further exemplified that the proposed appearance model and joint training are crucial to synthesize realistic examples compared to existing medical registration methods. The code, data, and models are available at \url{https://github.com/devavratTomar/SST/}.

 
\end{abstract}

\section{Introduction}\label{introduction}
Automated medical image segmentation, for example, to localize anatomical structures, is of great importance for disease diagnosis or treatment planning. Fully supervised deep neural networks (DNNs)  \cite{milletari2016v, ronneberger2015u} achieve state-of-the-art results when trained on large amounts of labeled data. However, acquiring abundant labeled data is often not feasible, as manual labeling is tedious and costly. Using available open-source data for domain adaptation \cite{tomar2021self,bozorgtabar2019using,bozorgtabar2019learn} or hand-tuned approaches for augmentation only provides suboptimal solutions. Furthermore, the cross-modality adaptation methods \cite{tomar2021self,tomar2021content,bozorgtabar2019syndemo} usually rely on fully labeled source datasets. Medical images can vary significantly from institution to institution in terms of vendor and acquisition protocols \cite{leung2010robust}. As a result, the pre-trained deep networks often fail to generalize to new test data that are not distributed identically to the training data. Furthermore, although hand-tuned data augmentation (DA) techniques \cite{akkus2017deep, moeskops2016automatic} caused improvement in segmentation accuracy \cite{pereira2016brain} partially, manual refinements are not sustainable, and augmented images have limited capacities to cover real variations and complex structural differences found in different images. Therefore, few-shot learning \cite{ouyang2020self, snell2017prototypical} or self-supervised learning \cite{chen2020simple, he2020momentum} based approaches that alleviate the need for large labeled data would be of crucial importance. However, these approaches have not been explored much for low labeled data regime medical image segmentation, e.g., one-shot scenarios. Another classical medical imaging approach often used to reduce the need for labeled data for synthesis and segmentation purposes is an atlas-based approach \cite{catana2010toward, arabi2016atlas, lorenzo2002atlas}. In this approach, an atlas is used to register each image and warp one into another (labels undergo the same transformation). One atlas (with its label) is enough for the procedure; however, one can utilize more if available to improve accuracy. Nevertheless, heterogeneity of medical images causes inaccurate warping and, consequently, erroneous segmentation. This heterogeneity issue is even more pronounced for the multi-site medical dataset. Recently, atlas-based approaches empowered by deep convolutional neural networks (CNNs) \cite{zhao2019data, elmahdy2019adversarial, yang2018neural} enable the development of one-shot learning segmentation models. 

Among recent one-shot learning methods, two are the most relevant to ours \cite{zhao2019data, wang2020lt}. In the first work, Zhao \textit{et al}. \cite{zhao2019data} proposed a learning framework to synthesize labeled samples using CNN to warp images based on atlas. However, \cite{zhao2019data} only recreates exact styles/deformations presented in the dataset \textbf{without inducing diversity}. Furthermore,  training includes two separate steps for spatial and appearance transformations bringing extra computational overhead. In the second work \cite{wang2020lt}, a one-shot segmentation method has been proposed based on the forward-backward consistency strategy to stabilize the training process. Nonetheless, since the atlas' style does not match the unlabeled image style, this results in imprecise registration and imperfect segmentation. In summary, our main contributions are as follows:

 \begin{itemize}
 
 \item We propose a novel volumetric self-supervised contrastive learning to learn style representation from unlabeled data facilitating registration and consequently segmentation task in the presence of intra-site and inter-site heterogeneity of MR images (see Fig.~\ref{fig:eye_catching});

  \item Unlike current state-of-the-art methods, our method does not require input volumes at inference time to synthesize new images. Instead, it can generate diversified and unlimited image-segmentation pairs by just sampling from a prior distribution; 

  \item Previous methods train the spatial and appearance models separately, resulting in sub-optimal compared to our joint optimization of all modules. Our method achieved state-of-the-art one-shot segmentation performance on two brain T1-weighted MRI datasets and improved the generalization ability for cross-site adaptation scenarios.
 
  
 
 

 \end{itemize}

\section{Related Work}\label{related_work}
\begin{figure*}[t!]
    \centering
    \vspace{-6mm}
    \includegraphics[width=\linewidth]{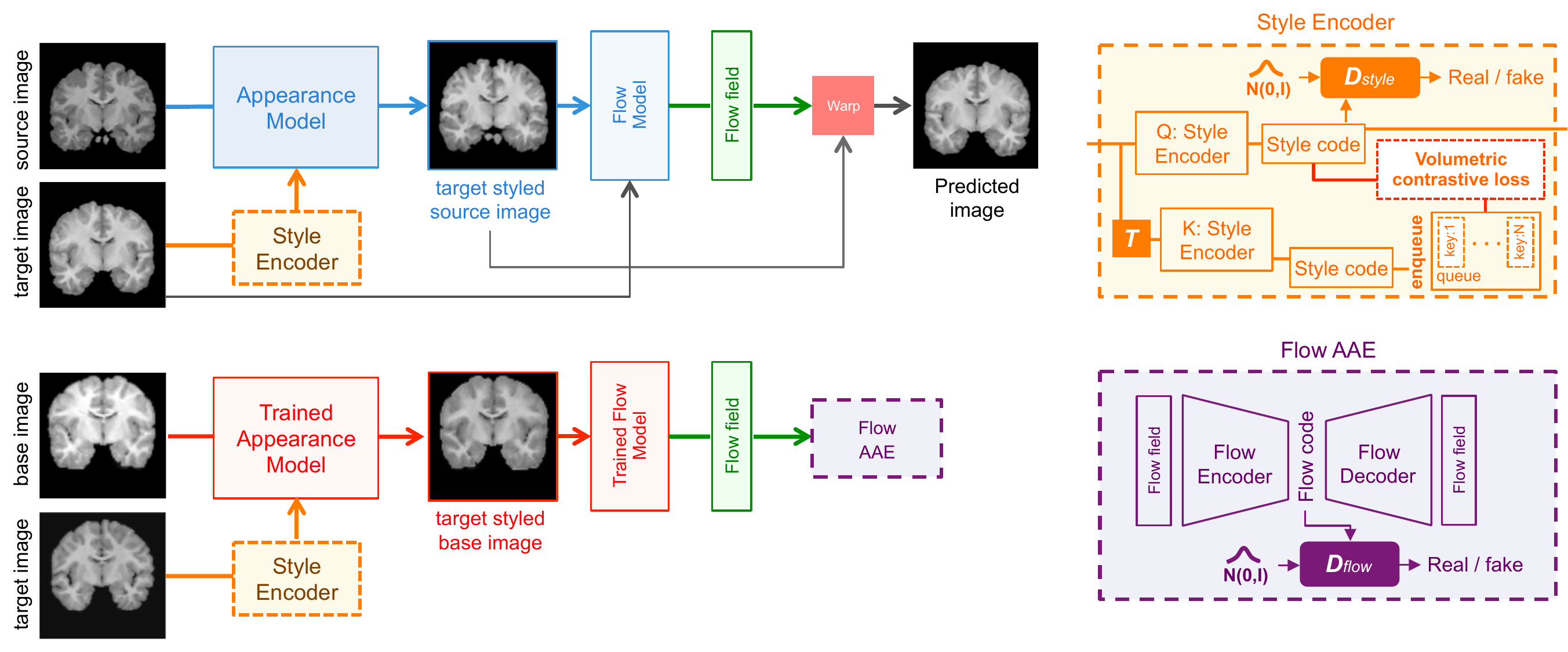}
    \caption{\textbf{Schematic description of the training phase.} The \textbf{Appearance Model} applies the target image's style to the source image using its style code predicted by the \textbf{Style Encoder}, which is trained  in parallel using self-supervised volumetric contrastive loss. Then, the \textbf{Flow Model} non-linearly warps the style translated source image to the target image. This allows backpropagation of supervision signals to all three models. Independently, \textbf{Flow AAE} maps the flow fields corresponding to the base image generated by using the trained \textbf{Flow model} and \textbf{Appearance model} to a normally distributed latent space. $\mathbf{D_\text{style}}$ and $\mathbf{D_\text{flow}}$ are trained in an adversarial manner to ensure that the style codes and the flow codes are normally distributed.}
    \vspace{-3mm}
    \label{fig:overview_proposed}
\end{figure*}


\textbf{Data Augmentation.}
Various data augmentation techniques have been developed to compensate for the extreme scarcity of labeled data encountered in medical image segmentation \cite{chen2020realistic,pre_olut2020_eccv,zhao2019data}. Augmentation approaches range from geometric transformations \cite{milletari2016v,oliveira2017augmenting} to data-driven augmentation methods \cite{han2020infinite, mahapatra2020pathological}. While former methods are often difficult to tune as they have limited capacities to cover real variations and complex structural differences found in images, latter approaches often learn generative models to synthesize images to combine with real images and train the segmentation model. However, variations in shape and anatomical structures can negatively impact their performance, especially when little training data is available. In this regard, image registration has been effective in approximating deformations between unlabeled images so that the augmented images with plausible deformations can be obtained \cite{zhao2019data,xu2019deepatlas,chaitanya2019semi}. The augmented images then allow training deep segmentation models with few labeled examples in a semi-supervised manner. Disappointingly, heterogeneity of medical images often yields inaccurate warping between the moving image and the fixed image and, consequently, inaccurate segmentation. Olut \textit{et al}. \cite{pre_olut2020_eccv} proposed an augmentation method to leverage statistical deformation models from unlabeled data via principal component analysis. Shen \textit{et al}. \cite{shen2020anatomical} proposed a geometric based image augmentation method that generates realistic images via interpolation from the geodesic subspace to estimate the space of anatomical variability. He \textit{et al}. \cite{he2020deep} proposed a Deep Complementary Joint Model (DeepRS) for medical image registration and segmentation.


\textbf{Image Segmentation in Low Data Regimes.}
Self-supervised learning and few-shot learning are two facets of the same problem: \textit{training a deep model in low labeled data regime}. These approaches have been used for sparsely annotated images for segmentation but without much success. The former often rely on many training classes to avoid overfitting \cite{wang2019panet,snell2017prototypical}, while the latter requires fine-tuning on sufficient labeled data before testing \cite{grill2020bootstrap,wang2020self}. Deep atlas-based models \cite{balakrishnan2019voxelmorph,zhao2019data,wang2020lt, dalca2019unsupervised, arabi2016atlas, lorenzo2002atlas} using a single atlas or multi-atlas tackled weakly-supervised medical image segmentation. Balakrishnan \textit{et al.} \cite{balakrishnan2019voxelmorph} developed VoxelMorph, which aims to estimate pairwise 3D image registration through a learned CNN-based spatial transformation function. In a one-shot learning context, it learns to register a labeled atlas to any other unlabeled volume. This model suffers from variance in voxel intensity confusing the spatial transformer. In this regard, a similar unsupervised method \cite{dalca2019unsupervised} has been proposed that combines a conventional probabilistic atlas-based segmentation with deep learning for MRI scan segmentation. More recently, few deep models \cite{zhao2019data,wang2020lt} explore the one-shot setting for medical image segmentation. Nevertheless, these methods either use samples from a single site (hospital) \cite{wang2020lt} or aggregate data from multiple sites \cite{zhao2019data} without cross-dataset transfer learning capability.

\section{Method}\label{method}
We first recap the concept of our proposed one-shot atlas-based medical image segmentation. We formulate the synthesis of novel volumetric images and their corresponding segmentation labels as a learned random spatial and style deformation of the given single labeled volumetric atlas image (referred to as \textbf{base image}) from learned latent space. We employ a \textbf{Style Encoder} (Sec.~\ref{style_encoder}) that learns to cluster similar styled images together in a self-supervised manner using volumetric contrastive loss by adapting Momentum Contrast \cite{he2020momentum} while imposing a normal distribution prior on the latent style codes using adversarial training \cite{makhzani2016adversarial}. The \textbf{Appearance Model} (Sec.~\ref{appearance_model}) is trained to generate different styles of the base image without changing its spatial structure. For learning the spatial deformation correspondences (referred to as \textbf{flow}) between the base image and the target image, we employ  \textbf{Flow Model} (Sec.~\ref{flow_model}) that is trained on the task of registering two different image volumes with the same style. This is achieved by first changing the style of moving image (referred to as \textbf{source image}) to match the target image's style (as obtained by \textbf{Style Encoder}) using the Appearance Model, followed by morphing it into the target image. As discussed later in Sec~\ref{ablation}, matching the source image's style to the target image improves the registration accuracy. We employ an additional adversarial autoencoder \cite{makhzani2016adversarial} that encodes the Flow model's output for the base image into a Gaussian prior flow latent space to learn the distribution of the spatial deformation fields corresponding to the base image. At test time, we sample flow latent code and style latent code from Gaussian prior to generate new deformation fields and style appearances for the base image, respectively. Fig.~\ref{fig:overview_proposed} and Fig.~\ref{fig:data_generation} show an overview of the training procedure and the data generation at test time, respectively. To quantify the images' quality and their corresponding segmentation labels generated using our approach, we train a separate 3D U-Net \cite{cciccek20163d} on them. We test our model on the real image/segmentation pairs and compare it with the performance of a fully supervised model trained using real data. The loss terms used for training Style Encoder, Appearance Model, and Flow Model, described in the subsequent subsections.

\subsection{Style Encoder}\label{style_encoder}
The Style Encoder aims to incentivize content-invariant image-level representation that brings together similar styled images and pushes apart dissimilar styled images. To do so, we propose a new volumetric contrastive learning \cite{chen2020simple,he2020momentum,chen2020improved} based strategy for training. In particular, we adapt Momentum Contrast \cite{he2020momentum} for volumetric medical images for clustering task of images' styles as opposed to the original classification task \cite{he2020momentum}. More importantly, we use the learned spatial transformer\footnote{The spatial transformer is trained as in \cite{balakrishnan2019voxelmorph}.} to generate the positive keys (preserving the styles) instead of standard augmentation, e.g., random cropping used in the original formulation. Without loss of generality, we keep a dictionary of keys $\{k_1, k_2, ..., k_K\}$ that represents different styles. During a training step, we sample a volumetric image $\mathbf{q}$ (called \textbf{query}) from the training set and generate its corresponding positive \textbf{key} volumetric image $\mathbf{k_+}$ by warping $\mathbf{q}$ to a randomly selected volumetric image from the training set using a pre-trained spatial transformer $\mathbf{\mathcal{T}}$. This ensures that $\mathbf{q}$ and $\mathbf{k_+}$ have the same style (with different structural geometries) that is different from the style keys in the dictionary. The dictionary's style keys are generated by a separate model (\textbf{key-Style Encoder}) whose weights are updated as a moving average of the weights of the Style Encoder with momentum $m=0.99$. The volumetric contrastive loss is computed as below:
\begin{equation}
    \mathcal{L}_{\text{vol\text{\_cl}}} = -\log\frac{\exp(q.k_+/\tau)}{\sum_{i=0}^{K}\exp(q.k_i/\tau)}
\end{equation}
where $k_+ = \mathcal{T}(q)$ and $\tau$ is a temperature hyper-parameter \cite{wu2018unsupervised}. The sum is over one positive and $K$ negative samples.

\subsection{Appearance Model}\label{appearance_model}
The Appearance Model is responsible for translating the style of the source image ($s$) to that of the target image ($t$), given the style latent code as predicted by the Style Encoder. We feed the target image's style code to the adaptive instance normalization (AdaIN) layers \cite{huang2017arbitrary} of the model to perform style transfer. Thus, the target styled source image ($\Tilde{s}$) is obtained as: 

\begin{equation}
    \Tilde{s} = \mathcal{A}(s, E_\text{style}(t))
\end{equation}
where $\mathcal{A}$ denotes the Appearance model, and $E_\text{style}$ represents the Style Encoder. The Appearance Model loss $\mathcal{L}_\text{app}$ consists of two components: $\mathcal{L}_\text{app} =\mathcal{L}^{\text{style}}_\text{cycle}+ \mathcal{L}^\text{style}_\text{id}$, where $\mathcal{L}^{\text{style}}_\text{cycle}$ and $\mathcal{L}^\text{style}_\text{id}$ denote the style consistency loss and style identity loss, respectively that are described below.

\paragraph{Style Consistency Loss.}
We include a style consistency loss that guides the Appearance model to generate images with the same spatial structure as the source image but with a different style in a consistent cyclic manner. Given the style codes of the source and target images, the following style consistency loss is computed as:

\begin{equation}
    \mathcal{L}^{\text{style}}_\text{cycle} = \mathcal{L}_{\text{SSIM-L}_1}(s, \mathcal{A}(\Tilde{s}, E_\text{style}(s)))
\end{equation}
\noindent
where $\mathcal{L}_{\text{SSIM-L}_1}$ computes multi-scale structural similarity index \cite{wang2003multiscale} and $L_1$ distance between the two images as:

\begin{equation}
    \mathcal{L}_{\text{SSIM-L}_1}(u, v) = \big\|u - v\big\|_1 + (1 -\text{SSIM}(u, v))
\end{equation}

\paragraph{Style Identity Loss.}
We also include a regularization loss term, called style identity, that enforces the Appearance Model to generate the same image using its own style.

\begin{equation}
\mathcal{L}^\text{style}_\text{id} = \mathcal{L}_{\text{SSIM-L}_1}(s, \mathcal{A}(s, E_\text{style}(s)))
\end{equation}

\subsection{Flow Model}\label{flow_model}
The Flow Model builds upon a spatial transformer network that warps a \textit{moving image} ($I_m$) to the \textit{fixed image} ($I_f$). The spatial transformer ($\mathcal{F}$) generates a correspondence map $\delta p$, referred to as \textit{flow}, which is used to register $I_m$ onto $I_f$. This warping operation is defined as:

\begin{equation}
\begin{split}
    \delta p &= \mathcal{F}(I_m, I_f) \\
    y &= \delta p \odot I_m
\end{split}
\end{equation}
where $\odot$ denotes the warping operator, and $y$ is the predicted image. Once we know the correspondence map $\delta p$ between the base image ($b$) and the target image ($t$), we can transfer the segmentation label of the base image ($b_\text{seg}$) onto the target image using the same warping operation.

\begin{equation}
    t_\text{seg} = \mathcal{F}(b, t)\odot b_\text{seg}
\end{equation}
To learn the distribution of deformation fields corresponding to the base image predicted by the Flow Model, we train a separate adversarial autoencoder \cite{makhzani2016adversarial} on its output. This allows us to encode the flow $\delta p$ in the latent space, which can be used later to generate novel volumetric images and corresponding segmentation labels (Sec.~\ref{generation_phase}). The Flow Model loss $\mathcal{L}_\text{flow}$ consists of two components: $\mathcal{L}_\text{flow} =\mathcal{L}^\text{flow}_\text{recon}+ \lambda_{\text{reg}}\mathcal{L}^\text{flow}_\text{reg}$, where $\mathcal{L}^\text{flow}_\text{recon}$ and $\mathcal{L}^\text{flow}_\text{reg}$ denote the reconstruction loss and the flow regularization loss that are described below.

\paragraph{Reconstruction Loss.}
In contrast to the Normalized Cross-Correlation loss \cite{balakrishnan2019voxelmorph} commonly used for the voxel registration, we include pixel similarity based reconstruction loss between the target image ($t$) and the spatially warped target styled source image (referred to as \textbf{predicted image}) obtained by Flow Model. Using the pixel-wise similarity loss is justified as the Appearance Model changes the source image's style to match the style of the target image, thus allowing adequate registration of the two images.

\begin{equation}
    \mathcal{L}^\text{flow}_\text{recon} = \mathcal{L}_{\text{SSIM-L}_1}(t, \mathcal{F}(\Tilde{s}, t) \odot \Tilde{s})
\end{equation}

\paragraph{Flow Regularization.}
We also regularize the flow $\delta p$ by penalizing its spatial gradient, thus ensuring the smoothness of the correspondence map generated by our spatial transformer.

\begin{equation}
    \mathcal{L}^\text{flow}_\text{reg} = \big\|\nabla\mathcal{F}(\Tilde{s}, t)\big\|_1
\end{equation}
We prefer the $L_{1}$ norm as it helps to stabilize training and results in less noisy flow compared to the $L_{2}$ norm.


\subsection{Adversarial Loss}
We introduce two latent discriminators, called $D_\text{style}$ and $D_\text{flow}$, for enforcing a prior distribution on the latent style codes and the flow codes generated during training, respectively. We use the
adversarial loss function described in LS-GAN \cite{mao2017squares} for training the two discriminators along with Style Encoder and Flow Encoder in an adversarial manner, as shown in Fig.~\ref{fig:overview_proposed}.

\begin{multline}
    \mathcal{L}^\text{style}_\text{adv} = \mathbb{E}_{t\sim X_\text{data}}\big[\big(D_\text{style}(E_\text{style}(t)) - 1\big)^2\big]\\ + \mathbb{E}_{n\sim \mathcal{N}}\big[D_\text{style}(n)^2\big]
\end{multline}
where $t$ is sampled from the training images $X_\text{data}$, and $\mathcal{N}$ is the normal distribution. Similarly, we train the adversarial autoencoder (AAE) \cite{makhzani2016adversarial} on the flow fields generated by Flow Model (Sec.~\ref{flow_model}) using LS-GAN loss and $l_1$ reconstruction loss. The optimization details of AAE are included in the \textbf{Appendix}.
\subsection{Training Objective}
Finally, the proposed training loss $\mathbf{\mathcal{L}_{total}}$ joins the loss terms used to train Style Encoder, Appearance Model, and Flow Model: 
\begin{equation}
\mathbf{\mathcal{L}_{total}} = \mathcal{L}_{\text{vol}\_\text{cl}} + \lambda_{1}\mathcal{L}_\text{app}+\lambda_{2} \mathcal{L}_\text{flow}+ \lambda_{3}\mathcal{L}^\text{style}_\text{adv}   
\end{equation}
where $\lambda$'s are the weights of different losses. We observed that pre-training the Style Encoder alone using the loss: $\mathcal{L}_{\text{vol\_cl}}+\lambda_3\mathcal{L}_\text{adv}^\text{style}$ improves the overall convergence rate and reduces the optimization's complexity. After pre-training the Style Encoder, we jointly optimize it along with Appearance Model and Flow Model by minimizing $\mathbf{\mathcal{L}_{total}}$. A sensitivity test is included on different values of $\lambda$'s in the \textbf{Appendix}.

\begin{figure}
  \includegraphics[width=\linewidth]{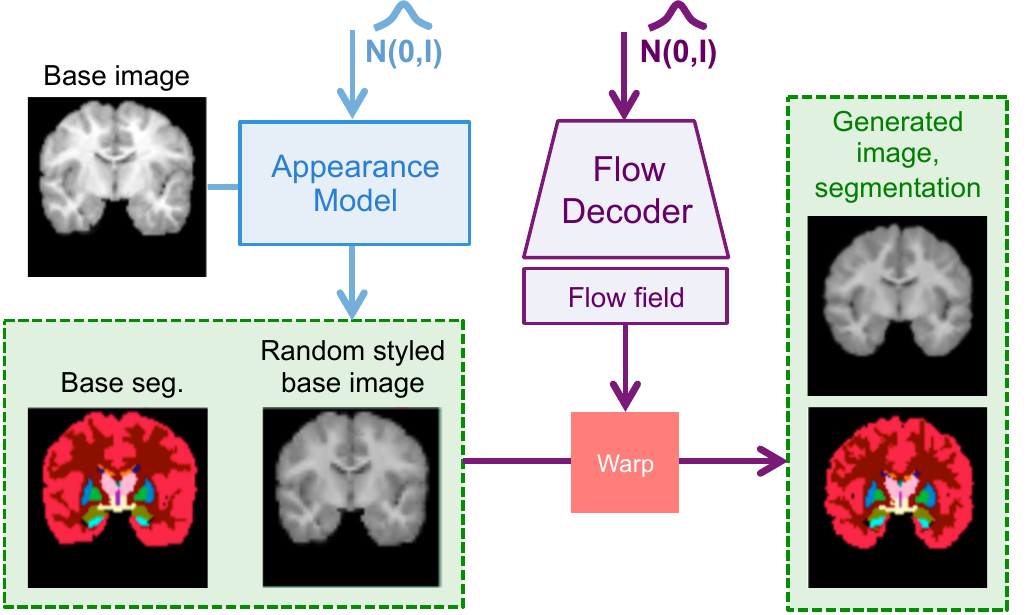} 
  \caption{\textbf{Data Generation} phase. We use the learned \textbf{Appearance Model} and \textbf{Flow Decoder} to generate new images and segmentation labels from the normal distribution.}
  \label{fig:data_generation}
\end{figure}

\subsection{Generation Phase}\label{generation_phase}
As shown in Fig.~\ref{fig:data_generation}, we can sample new volumetric images and their segmentation maps by mapping Gaussian noise to novel style codes and flow fields using the trained Appearance Model ($\mathcal{A}$) and Flow Decoder ($G_\text{flow}$). The Appearance Model performs style deformation of the base image. At the same time, Flow Decoder produces a random flow field, which is then used to warp the styled base image and the base segmentation, thus generating new image-segmentation pairs.

\begin{align}
    X &= G_\text{flow}(n_\text{flow}) \odot \mathcal{A}(b, n_\text{style})\\
    Y &= G_\text{flow}(n_\text{flow}) \odot b_\text{seg}\nonumber
\end{align}
where $b$ and $b_\text{seg}$ are the base image and base segmentation and $n_\text{flow}$, $n_\text{style}$ $\sim$ $\mathcal{N}$. $X$ and $Y$ are the generated image-segmentation pairs.

\section{Experiments}\label{experiments}
This section introduces the implementation details, experimental settings, dataset, results, and ablation studies.
\begin{table*}[t]
\centering
\resizebox{\textwidth}{!}{
\begin{tabular}{@{}lccccccccccccccccc@{}}
                    & \rotatebox{80}{Cerebral-WM}  & \rotatebox{80}{Cerebral-CX} & \rotatebox{80}{Lateral-Vent} & \rotatebox{80}{Cerebellum-WM}& \rotatebox{80}{Cerebellum-CX} & \rotatebox{80}{Thalamus-Proper} & \rotatebox{80}{Caudate} & \rotatebox{80}{Putamen} & \rotatebox{80}{Pallidum} & \rotatebox{80}{3rd-Vent} & \rotatebox{80}{4th-Vent} & \rotatebox{80}{Brain-Stem} & \rotatebox{80}{Hippocampus} & \rotatebox{80}{Amygdala}& \rotatebox{80}{CSF} & \rotatebox{80}{VentralDC} & Mean\std{std}\\ \hline

\multicolumn{18}{c}{\cellcolor[HTML]{EFEFEF}\textbf{(CANDI)}}\\ \hline
MABMIS & $86.3$ & $90.5$ & $86.1$ & $75.9$ & $90.2$ & $84.2$ & $79.2$ & $79.5$ & $67.6$ & $66.7$ & $79.8$ & $90.8$ & $73.4$ & $64.8$ & $60.4$ & $77.0$ & $78.3$\std{2.9}\\
VoxelMorph &   $81.1$ & $87.3$ & $83.7$ & $69.9$ & $82.4$ & $85.9$ & $82.6$ & $81.0$ & $74.7$ & $64.3$ & $73.4$ & $89.2$ & $66.2$ & $66.7$ & $59.4$ & $81.1$ & $76.9$\std{3.2} \\
Bayesian & $89.5$ & $84.5$ & $85.3$ & $\textcolor{onlineBlue}{\mathbf{84.9}}$ & $82.4$ & $82.1$ & $83.7$ & $83.0$ &      
$77.2$ & $57.5$ & $75.7$ & $84.1$ & $74.3$ & $\textcolor{onlineBlue}{\mathbf{72.8}}$ & $48.6$ & $75.9$ & $77.7$\std{2.3}     \\
DataAug & $84.8$ & $89.0$ & $77.4$ & $72.3$ & $86.8$ &	
$89.0$ & $84.6$ & $86.3$ & $79.8$ &	$71.2$ & $78.7$ & $91.3$ & $\textcolor{onlineBlue}{\mathbf{76.0}}$ & $72.3$ & $63.3$ & $\textcolor{onlineBlue}{\mathbf{82.7}}$ & $80.4$\std{3.2} \\
LT-Net** & $85.8$  & $90.9$ & $83.1$  & $80.1$ & $\textcolor{onlineBlue}{\mathbf{91.6}}$ & $87.9$ & $85.5$ & $\textcolor{onlineBlue}{\mathbf{88.4}}$  & $80.5$ & $68.4$ & $79.7$ & $\textcolor{onlineBlue}{\mathbf{92.4}}$ & $71.6$ & $71.6$ & $67.1$ & $82.3$  & $81.7$\std{8.0} \\
\textbf{Ours}  & $\textcolor{onlineBlue}{\mathbf{90.9}}$ & $\textcolor{onlineBlue}{\mathbf{94.3}}$ & $\textcolor{onlineBlue}{\mathbf{89.2}}$ & $83.4$ & $89.5$  & 
$\textcolor{onlineBlue}{\mathbf{89.2}}$ & $\textcolor{onlineBlue}{\mathbf{88.3}}$ & $86.7$ & $\textcolor{onlineBlue}{\mathbf{81.1}}$ & $\textcolor{onlineBlue}{\mathbf{71.3}}$ & $\textcolor{onlineBlue}{\mathbf{81.9}}$ & $92.2$ & $\textcolor{onlineBlue}{\mathbf{76.0}}$ & $70.9$ & $\textcolor{onlineBlue}{\mathbf{67.6}}$ & $82.2$ & $\textcolor{onlineBlue}{\mathbf{83.5}}$\std{3.0} \\
\hline
3D U-Net* & $\mathbf{94.1}$  & $\mathbf{97.0}$ & $\mathbf{93.8}$ & $\mathbf{89.6}$ & $\mathbf{96.8}$ & $\mathbf{91.5}$ & $\mathbf{90.4}$ & $\mathbf{90.8}$ & $\mathbf{83.0}$ & $\mathbf{74.4}$ & $\mathbf{87.0}$ & $\mathbf{94.9}$ & $\mathbf{84.1}$  & $\mathbf{79.0}$    & $\mathbf{79.6}$  & $\mathbf{88.1}$ & $\mathbf{88.1}$\std{\textbf{1.5}} \\ 
\bottomrule
\multicolumn{18}{c}{\cellcolor[HTML]{EFEFEF}\textbf{(OASIS)}}\\ \hline

MABMIS & $79.4$ & $62.1$ & $84.4$ & $77.1$ & $82.6$ & $\textcolor{onlineBlue}{\mathbf{85.3}}$ & $77.5$ & $78.9$ & $66.5$ & $81.5$ & $62.3$ & $88.4$ & $72.5$ & $72.9$ & $63.6$ & $78.3$ & $75.8$\std{6.9}\\
VoxelMorph &   $75.2$ & $63.1$ & $85.3$ & $77.9$ & $83.4$ & $85.0$ & $79.1$ & $\textcolor{onlineBlue}{\mathbf{81.3}}$ & $\textcolor{onlineBlue}{\mathbf{72.3}}$ & $81.0$ & $58.8$ & $90.5$ & $70.3$ & $\textcolor{onlineBlue}{\mathbf{72.8}}$ & $\textcolor{onlineBlue}{\mathbf{67.6}}$ & $\textcolor{onlineBlue}{\mathbf{79.4}}$ & $76.4$\std{4.3} \\
Bayesian & $\textcolor{onlineBlue}{\mathbf{90.0}}$ & $45.8$ & $64.7$ & $82.4$ & $72.9$ & $84.1$ &      
$70.1$ & $70.7$ & $40.6$ & $33.4$ & $18.8$ & $87.8$ & $48.3$ & $56.9$ & $33.0$ & $62.1$ & $59.5$\std{3.0}     \\
DataAug & $74.8$ & $69.0$ & $69.2$ & $67.2$ & $80.1$ &	
$78.0$ & $61.8$ &	$66.2$ & $50.4$ & $70.5$  & $50.6$ & $83.2$ & $64.8$ & $56.8$ & $56.8$ & $69.2$ & $66.8$\std{5.7} \\
\textbf{Ours}  & $89.4$ & $\textcolor{onlineBlue}{\mathbf{89.2}}$ & $\textcolor{onlineBlue}{\mathbf{89.2}}$ & $\textcolor{onlineBlue}{\mathbf{86.3}}$ & $\textcolor{onlineBlue}{\mathbf{91.7}}$  & $84.8$ & $\textcolor{onlineBlue}{\mathbf{80.5}}$ & $80.1$ & $65.1$ & $\textcolor{onlineBlue}{\mathbf{82.0}}$ & $\textcolor{onlineBlue}{\mathbf{70.3}}$ & $\textcolor{onlineBlue}{\mathbf{91.5}}$ & $\textcolor{onlineBlue}{\mathbf{74.7}}$ & $69.4$ & $65.4$ & $77.9$ & $\textcolor{onlineBlue}{\mathbf{80.5}}$\std{3.9} \\
\hline
3D U-Net* & $\mathbf{94.2}$  & $\mathbf{95.6}$ & $\mathbf{94.6}$ & $\mathbf{92.0}$ & $\mathbf{97.5}$ & $\mathbf{91.8}$ & $\mathbf{89.1}$ & $\mathbf{89.5}$ & $\mathbf{79.7}$ & $\mathbf{90.9}$ & $\mathbf{89.8}$  & $\mathbf{96.9}$ &  $\mathbf{89.3}$ & $\mathbf{85.1}$ & $\mathbf{86.3}$  & $\mathbf{87.9}$ & $\mathbf{90.6}$\std{\textbf{1.4}}\\ 
\bottomrule
\multicolumn{18}{c}{\cellcolor[HTML]{EFEFEF}\textbf{(OASIS$\,\to\,$CANDI)}}\\
\hline
w/o Style Adap. & $84.9$ & $88.4$ & $58.9$ & $75.3$ & $90.6$ & $78.8$ & $55.5$ & $47.9$ & $40.6$ & $55.5$ & $65.5$ & $83.3$ & $55.2$ & $44.8$ & $44.6$ & $60.9$ & $64.4$\std{3.5}\\
w/ Style Adap. & $85.0$  & $90.0$ & $77.0$ & $74.7$ & $89.8$ & $83.3$ & $78.1$ & $75.8$ & $68.6$ & $66.3$ & $73.6$ & $88.1$ & $58.8$ & $52.3$ & $52.4$ & $69.2$ & $74.0$\std{3.2}\\

\bottomrule
\multicolumn{18}{c}{\cellcolor[HTML]{EFEFEF}\textbf{(CANDI$\,\to\,$OASIS)}}\\
\hline
w/o Style Adap. & $80.2$ & $83.1$ & $58.1$ & $32.4$ & $71.4$ & $37.2$ & $39.6$ & $20.6$ & $6.2$ & $3.8$ & $7.6$ & $12.5$ & $12.0$ & $17.9$ & $2.3$ & $12.6$ & $31.1$\std{11.7}\\
w/ Style Adap. & $85.4$ & $85.4$ & $81.6$ & $45.3$ & $67.8$ & $67.6$ & $51.9$ & $29.0$ & $6.6$ & $30.5$ & $14.2$ & $67.7$ & $31.8$ & $45.0$ & $32.2$ & $44.2$ & $49.1$\std{10.2}\\\hline
\end{tabular}
}
\caption{Comparison of segmentation performance (mean \textit{Dice} score in \%) of MABMIS (2 atlas) \cite{jia2012iterative}, VoxelMorph \cite{balakrishnan2019voxelmorph}, Bayesian \cite{dalca2019unsupervised}, DataAug \cite{zhao2019data}, LT-Net** \cite{wang2020lt}, and \textbf{Ours} across various brain structures on CANDI and OASIS datasets. Abbreviations used: white matter (WM), cortex (CX), ventricle (Vent), and cerebrospinal fluid (CSF). ** as reported in the published paper. * fully supervised model. The last four rows (OASIS$\to$CANDI and CANDI$\to$OASIS) indicate the results of our method with (w/) and without (w/o) cross-site style adaptation.}\label{tab:Dice_results_sota}
\end{table*}

\textbf{Dataset, Preprocessing and Evaluation Metric.}\label{dataset}
We evaluate our method and other baselines on multi-study datasets from publicly available datasets: CANDI \cite{kennedy2012candishare} and a large-scale dataset, OASIS \cite{marcus2007open}, each contains 3D T1-weighted MRI volumes. CANDI dataset consists of scans from 103 patients with manual anatomical segmentation labels, whereas the OASIS dataset has 2044 scans with segmentation labels obtained by the FreeSurfer \cite{fischl2012freesurfer} pipeline. As in VoxelMorph \cite{balakrishnan2019voxelmorph} and LT-Net \cite{wang2020lt}, 28 anatomical structures are used in our experiments. All the dataset images are first prepossessed by removing the brain's skull region, followed by center cropping the volumes to $128\times160\times160$. We set aside 20\% brain images and their segmentation as a test set, which was untouched during training. The remaining 80\% brain images are then used for training and validation, with a split of 90\%-10\%, in which there is no patient ID overlap among the subsets. For each dataset, there are different acquisition details and health conditions. The most similar image to the anatomical average is selected as the only annotated atlas (base image) used for the training. We only use validation set labels for choosing the best model and hyper-parameters. For the OASIS dataset, the model is trained with a mixture of healthy subjects and diseased patients and is then evaluated on test cases constitutes of both sets. We assess our method's efficiency by training a 3D U-Net-based segmentation model on the generated volumetric image-segmentation pairs and evaluating its performance on the untouched test data. We use the \textit{Dice} similarity coefficient between the ground truth segmentation and the predicted result in assessing the segmentation accuracy.

\begin{figure}[t]
    \centering
    \includegraphics[width=\columnwidth]{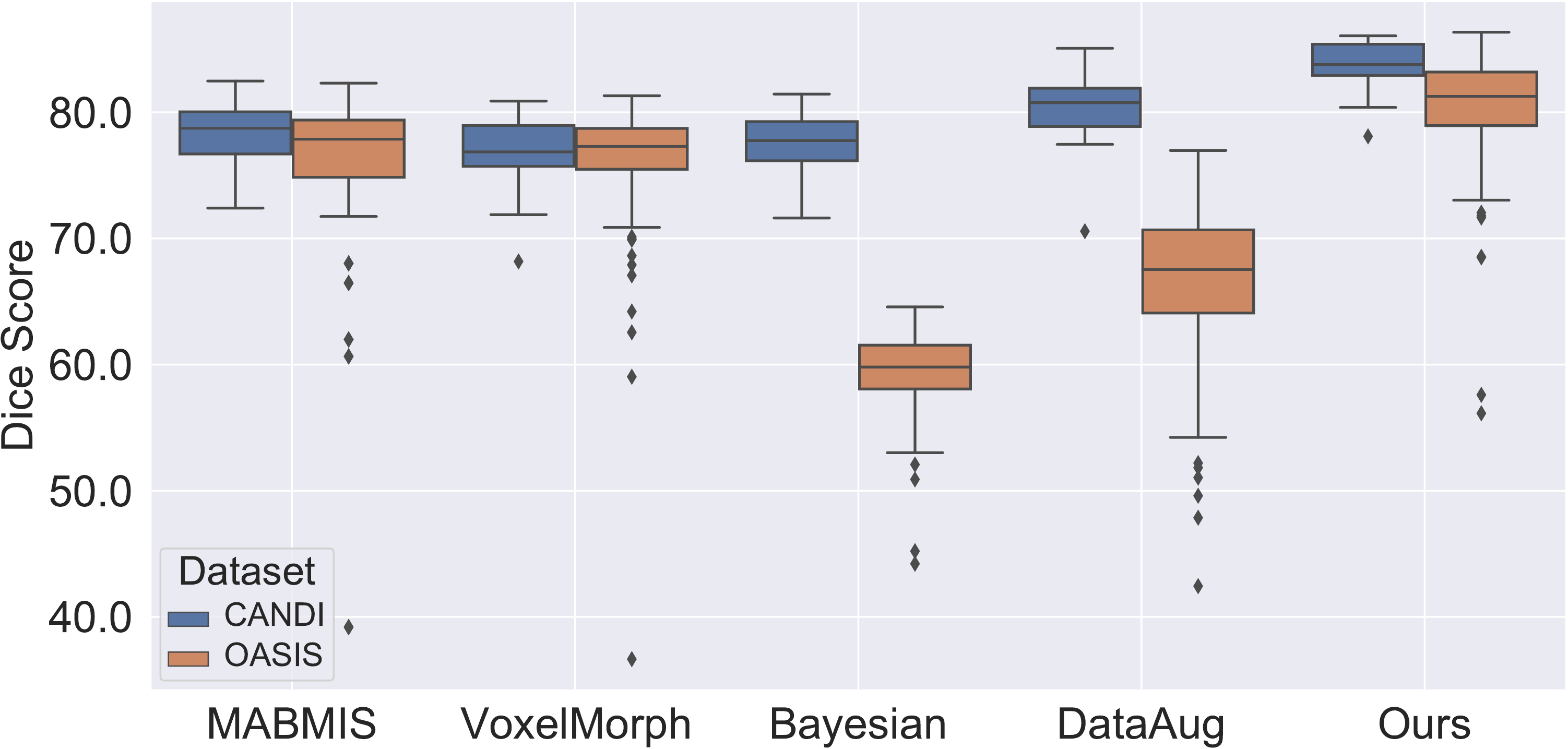}
    \caption{Comparison of volume-wise segmentation accuracy (\textit{Dice} score $\%$) of our method with MABMIS (2 atlas) \cite{jia2012iterative}, VoxelMorph \cite{balakrishnan2019voxelmorph}, Bayesian \cite{dalca2019unsupervised} and DataAug \cite{zhao2019data}. We outperform the second best baseline by a margin of $\mathbf{3.1}$\textbf{\footnotesize{\%}} on CANDI and $\mathbf{4.1}$\textbf{\footnotesize{\%}} on OASIS dataset (p-value of $\mathbf{5.8\times10^{-4}}$, $\mathbf{3.2\times10^{-16}}$ respectively using paired t-test).}
    \label{fig:distr_comparison}
\end{figure}

\begin{figure*}[!h]
    \centering
    \vspace{-3mm}
    \resizebox{\textwidth}{!}{%
    \begin{tabular}{@{}l@{}c@{}c@{}c@{}c@{}c@{}c@{}c@{}}
    \rotatebox{90}{\hspace{5em}CANDI} &\includegraphics[width=0.14\linewidth]{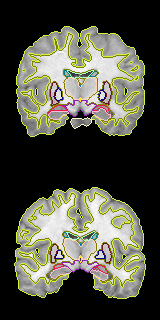} & \includegraphics[width=0.14\linewidth]{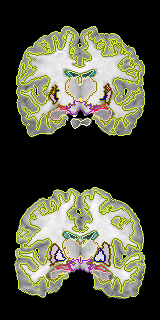} &
    \includegraphics[width=0.14\linewidth]{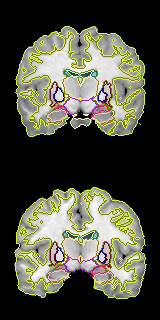} &
    \includegraphics[width=0.14\linewidth]{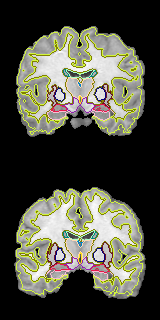} &
    \includegraphics[width=0.14\linewidth]{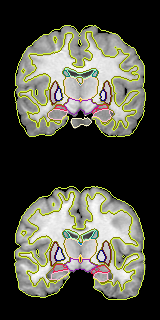} &
    \includegraphics[width=0.14\linewidth]{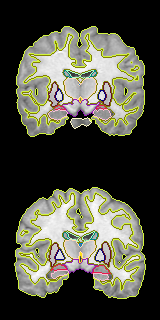} &
    \includegraphics[width=0.14\linewidth]{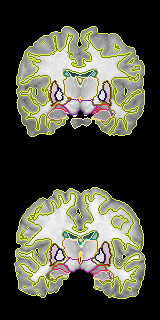}\\
    \rotatebox{90}{\hspace{5em}OASIS} &\includegraphics[width=0.14\linewidth]{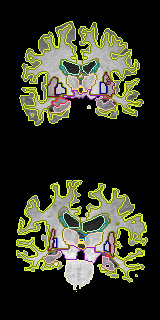} &
    \includegraphics[width=0.14\linewidth]{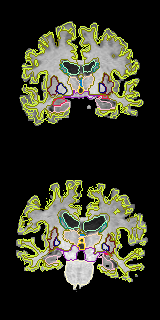} &
    \includegraphics[width=0.14\linewidth]{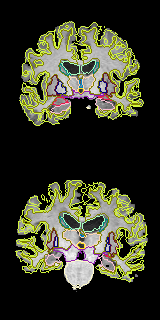} &
    \includegraphics[width=0.14\linewidth]{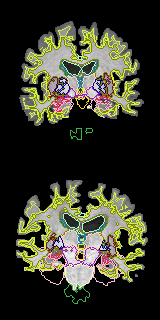} &
    \includegraphics[width=0.14\linewidth]{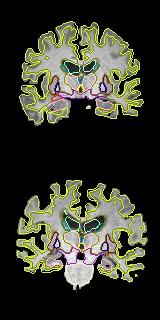} &
    \includegraphics[width=0.14\linewidth]{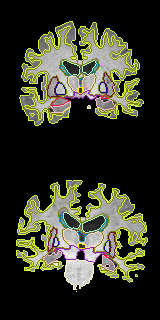} &
    \includegraphics[width=0.14\linewidth]{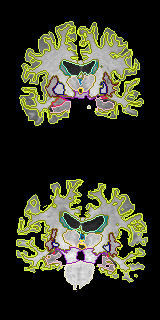}\\
    \hspace{1em} & (a) & (b) & (c) & (d) & (e) & (f) & (g)
    \end{tabular}
    }
    \caption{Qualitative comparison of our method with other baselines on the segmentation task of Brain MRI volumetric images from CANDI and OASIS dataset. Left to right: (a) Ground Truth, the segmentation results from (b) MABMIS \cite{jia2012iterative}, (c) VoxelMorph \cite{balakrishnan2019voxelmorph}, (d) Bayesian \cite{dalca2019unsupervised}, (e) DataAug \cite{zhao2019data}, (f) Fully Supervised 3D U-Net, (g) \textbf{Ours}. Best visualized in color.}
    \label{fig:quality_baseline_comparision}
\end{figure*}



\textbf{Experimental Settings.}
All our models are based on 3D CNNs \cite{ji20123d}. The Appearance model has $3$ AdaIN \cite{huang2017arbitrary} layers for performing style transfer using the style codes. Flow Model is a lighter version of VoxelMorph \cite{balakrishnan2019voxelmorph}, while AAE Model has a 3D convolutional encoder-decoder architecture. We implement all the models in PyTorch \cite{NEURIPS2019_9015}. The architectural details of all the models are included in the \textbf{Appendix}. For training Style Encoder, Appearance Model, Flow Model, and Flow Autoencoder, we used Adam \cite{kingma2014adam} optimizer with a learning rate of $2\times 10^{-4}$ and $\left ( \beta _{1}=0.9,\beta _{2}=0.999 \right )$. We use the same optimizer with the same learning rate for training the latent style code and flow code's discriminators but with $\left ( \beta _{1}=0.5,\beta _{2}=0.999 \right )$. A hyper-parameter search was conducted to find the optimal values. We choose the loss term weights as $\lambda_1=5.0$, $\lambda_2=1.0$, $\lambda_3=0.1$ and $\lambda_\text{reg}=0.1$. The temperature coefficient $\tau$ for volumetric contrastive loss is set to $0.7$. We use a batch size of $32$ for pre-training the Style Encoder and $4$ when all the models are trained end-to-end. We utilize the same experimental setup for all baseline experiments for a fair comparison, e.g., the same atlas.

\textbf{Comparison with SOTA Methods.}\label{sota}
Our method surpasses the state-of-the-art methods in most semantic classes and, on average, on both CANDI and OASIS datasets (see Fig.~\ref{fig:distr_comparison}). A qualitative comparison of our method with several baselines on coronal brain slices' segmentation task is shown in Fig.~\ref{fig:quality_baseline_comparision}.
MAMBIS~\cite{jia2012iterative} and VoxelMorph \cite{balakrishnan2019voxelmorph} produce visibly noisy and inaccurate boundaries compared to other methods. Bayesian \cite{dalca2019unsupervised} and DataAug~\cite{zhao2019data} struggle with outer contours, where one crops it unnecessarily, and another includes background. Furthermore, VoxelMorph and DataAug encounter difficulties in identifying small segmentation regions, whereas Bayesian overproduces them. Instead, our method handles outer/inner regions and smaller anatomical regions well and compares closely to supervised/ground truth results. The qualitative observations are backed by quantitative metrics (see Table \ref{tab:Dice_results_sota}). 


\subsection{Ablations}\label{ablation}

\begin{figure}
    \centering
    \includegraphics[width=\columnwidth]{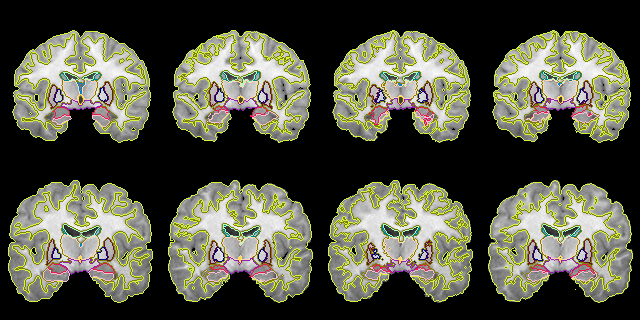}
    \caption{Qualitative comparison of registration accuracy evaluated on segmentation for different methods. \textit{Left to Right}: Ground-truth, VoxelMorph\cite{balakrishnan2019voxelmorph}, MABMIS\cite{jia2012iterative}, Ours.}
    \vspace{-2mm}
    \label{fig:registration_evaluation}
\end{figure}

\begin{figure}
    \centering
    \includegraphics[width=\columnwidth]{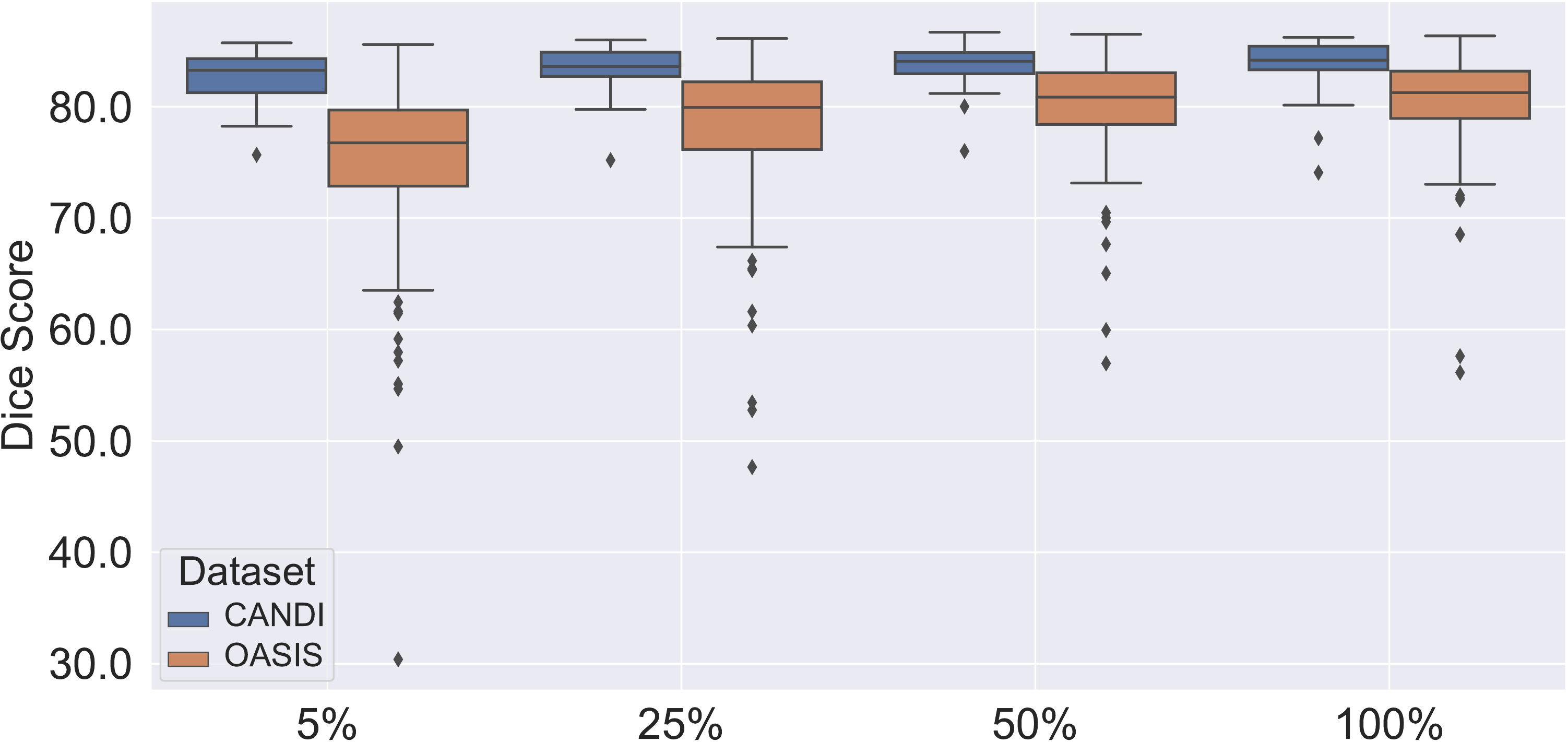}
    \caption{Ablation for segmentation accuracy of 3D U-Net using different sizes of generated data on CANDI and OASIS datasets. $100\%$ implies $1850$ generated image-segmentation volumes.}
    \label{fig:data_aug_diff_size}
\end{figure}
\begin{figure}[h]
\vspace{-1mm}
\begin{tabular}{@{}l@{}c@{}c@{}c@{}c@{}}
& \multicolumn{4}{l}{Different Flows $\longrightarrow$} \\
\rotatebox{90}{Different Styles $\longrightarrow$}   \includegraphics[width=0.23\linewidth]{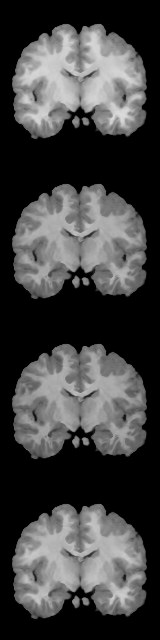} &  \includegraphics[width=0.23\linewidth]{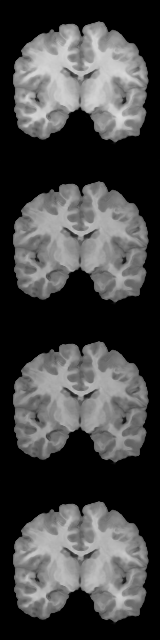}  & \includegraphics[width=0.23\linewidth]{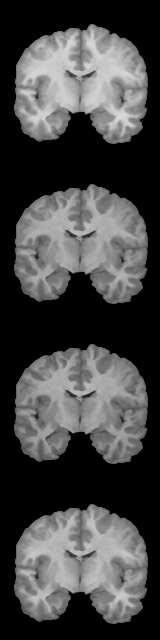}  & \includegraphics[width=0.23\linewidth]{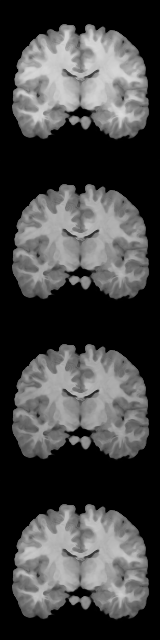}  & 
\end{tabular}
\caption{\textbf{Manipulating style and flow codes.} \textit{Left to right}: Images generated using the same style code with different flow codes. \textit{Bottom to top}: Images generated using the same flow code and distinct style.}
\label{fig:flow_style}
\end{figure}

\begin{table}
\centering
\begin{tabular}{@{}l|c|cc@{}}
\hline
\multirow{2}{*}{Ablation Type}                   & \multirow{2}{*}{Method} & \multicolumn{2}{c}{Mean\std{std}} \\ \cline{3-4}
                                            &                         & \multicolumn{1}{c|}{CANDI} & OASIS \\ \hline
w/o $\mathcal{L}_\text{app}$                & Reg. (a)      & \multicolumn{1}{c|}{$71.6$\std{$3.2$}} & $68.5$\std{$3.8$}\\
w/o Joint opt.                         & Reg. (a)        & \multicolumn{1}{c|}{$73.6$\std{$3.3$}} & $69.9$\std{$3.7$}\\
w/ Joint opt., $\mathcal{L}_\text{app}$ & Reg. (a)       & \multicolumn{1}{c|}{$\mathbf{78.9}$\std{$\mathbf{2.4}$}} & $\mathbf{73.3}$\std{$\mathbf{3.1}$}\\
w/o $\mathcal{L}_\text{app}$ & Sup. (b)  &
\multicolumn{1}{c|}{$54.9$\std{$20.4$}} & $41.2$\std{22.7}\\
w/ $\mathcal{L}_\text{app}$  & Sup. (b)  & \multicolumn{1}{c|}{$\mathbf{83.5}$\std{$\mathbf{3.0}$}} & $\mathbf{80.5}$\std{$\mathbf{3.9}$}\\ \hline
\end{tabular}
\caption{\textbf{Effect of style transfer and joint training.} Mean \textit{Dice} score of the segmentation task with and without Appearance Model and joint training using the method as (a) registration of the base image to the target image (b) 3D U-Net trained on generated images-segmentation pairs.}
\label{tab:ablation_appearance}
\end{table}


\textbf{Ablation Study on Appearance Model.}
The Appearance Model plays a crucial role in generating diversified styled images and improving the Flow Model's image registration efficiency during training. Table \ref{tab:ablation_appearance} shows the effect of including-excluding the Appearance Model ($\mathcal{L}_\text{app}$) on the registration accuracy of segmentation labels predicted by flow model and supervised segmentation accuracy of the 3D U-Net using the generated images. For both these scenarios, the average \textit{Dice} score improves after the Appearance Model's inclusion. This is expected as registering two images with similar intensity distributions is easier than registering images with different intensities. We can only generate similar styled images without the Appearance model for the generation phase, which leads to a poor generalization of 3D U-Net on the test set. A qualitative evaluation of registration on the CANDI test set is shown in Fig.~\ref{fig:registration_evaluation}.

\textbf{Ablation Study on Joint Training.}
We observed that training our model end-to-end is critical for improved registration accuracy (Table~\ref{tab:ablation_appearance}). We experimented with the pre-training Appearance model, Flow Model, and Style encoder separately using the losses defined in Sec.~\ref{method} followed by fine-tuning and found significant improvement when Appearance Model and Flow Model are trained end-to-end. However, a pre-training Style encoder using volumetric contrastive loss improves the overall model convergence ({$\mathbf{10\times}$} faster) without qualifying the performance as compared to training it jointly. 

%

\textbf{Cross-Site One-Shot Adaptation.}\label{multi-site}
To evaluate the Appearance model's efficacy in generating the unlabeled target dataset styles, we experimented with replacing the Appearance Model trained on CANDI (OASIS) with the one trained on OASIS (CANDI) to generate images in the style of the target domain. The volumetric image-segmentation pairs generated are then used to train 3D U-Net, which is further evaluated on the target domain's test samples. As shown in Table~\ref{tab:Dice_results_sota}, the domain gap between the two data sites would severely impede the generalization ability of the trained supervised model on the source data site (w/o style adaptation) while performing style adaptation between OASIS and CANDI dataset provides substantial improvements. We use the same test samples from CANDI and OASIS datasets for evaluation.

\textbf{Exploring Diversity of the Generated Data.}
We evaluate 3D U-Net's performance on the segmentation task using different data sizes generated by our approach. For the best performing 3D U-Net, we report the accuracy obtained by training it on $1850$ generated image-segmentation pairs. Fig.~\ref{fig:data_aug_diff_size} shows a box-plot of the \textit{Dice} score ($\%$) of the segmentation accuracy of 3D U-Net trained on different sizes of samples generated using our proposed method trained on CANDI and OASIS. This suggests that our method has learned to generate more diverse and realistic data. Besides, the effects of different flow codes and styles on generated images can be observed in Fig.~\ref{fig:flow_style}. 

\section{Conclusion and Future Work}\label{conclusion}
We proposed the novel volumetric contrastive loss used for style transfer by leveraging unlabeled data for one-shot medical image segmentation. We presented a generic method adapted for cross-site one-shot segmentation scenarios to generate arbitrarily diversified volumetric image-segmentation pairs using trained appearance models from one data site and a flow model from another data site. We demonstrated state-of-the-art one-shot segmentation performance on two T1-weighted brain MRI datasets under various settings and ablations. We shed light on our method’s efficacy in closing the gap with a fully-supervised segmentation model in the extreme case of only one labeled atlas. Our method uses neither tissue nor modality-specific information and can be adjusted to other modalities or anatomy. As future work, our method can be easily extended for the few-shot scenario using several atlases.

{\small
\bibliographystyle{ieee_fullname}
\bibliography{egbib}
}

\onecolumn


\section{Appendix}\label{Appendix}

\subsection{Sensitivity Test}\label{sensitivity}
We conduct a sensitivity test on our training objective's hyperparameters ($\lambda$'s) for CANDI and OASIS datasets in Table \ref{tab:sensitivity}. To be consistent, we chose the same $\lambda$'s values that are optimized for the OASIS dataset for the final quantitative evaluation.

\begin{table}[h!]
\centering
\begin{tabular}{@{}l|c|c|c|c|c|c|c|c@{}}
\hline
& Dataset & Method & $\lambda_1=1.0$ & $\lambda_1=10.0$ & $\lambda_2=2.0$ & $\lambda_2=5.0$ & $\lambda_3=0.2$ & $\lambda_3=0.5$\\ 
\hline
\multirow{4}{*}{Mean\std{std}} & \multirow{2}{*}{OASIS} & Reg. (a) & $69.7$\std{5.0} & $70.9$\std{5.6} & $71.6$\std{4.9} & $70.9$\std{5.1} & $72.5$\std{4.9} & $71.9$\std{5.0} \\
 & & Sup. (b) & $76.0$\std{4.1} & $77.5$\std{5.3} & $78.6$\std{4.3} & $76.1$\std{5.3} & $79.8$\std{4.6} & $77.6$\std{4.7} \\
 & \multirow{2}{*}{CANDI} & Reg. (a) & $77.7$\std{2.9} & $77.8$\std{3.0} & $77.6$\std{3.1} & $77.5$\std{3.2} & $78.9$\std{2.5} & $78.6$\std{2.6} \\
 & & Sup. (b) & $84.4$\std{2.0} & $85.7$\std{2.6} & $86.1$\std{2.0} & $85.5$\std{2.1} & $86.0$\std{2.1} & $84.8$\std{2.0} \\
\hline
\end{tabular}
\caption{\textbf{Sensitivity test for different values of $\lambda$'s for the proposed training loss.} We change one hyperparameter at a time while other hyperparameters' values are kept the same. Method Reg. (a) denotes registration accuracy of the base image and the target image; Method Sup. (b) denotes accuracy obtained by 3D U-Net trained on generated image-segmentation pairs.}
\label{tab:sensitivity}
\end{table}

\subsection{The Architectural Details}\label{architectures}
Throughout this appendix, we introduce several notations in the figures defined as:

\begin{itemize}
    \item \textbf{Conv3D} denotes a 3D convolution layer.
    \item \textbf{ConvT3D} denotes a 3D transpose convolution layer.
    \item \textbf{ks} denotes the size of the kernel.
    \item \textbf{stride} denotes the shift of the convolutional kernel in pixels while performing convolution.
    \item \textbf{nf} denotes the number of output features. 
    \item \textbf{Linear} denotes a fully connected multi-layer perceptron.
    \item \textbf{Upsample} denotes the upsampling of the features in the spatial dimensions.
    \item \textbf{Concat} denotes the concatenation of feature maps of two layers.
\end{itemize}

\subsubsection{Style Encoder's Architecture}
\begin{figure}[h]
    \centering
    \includegraphics[width=\linewidth]{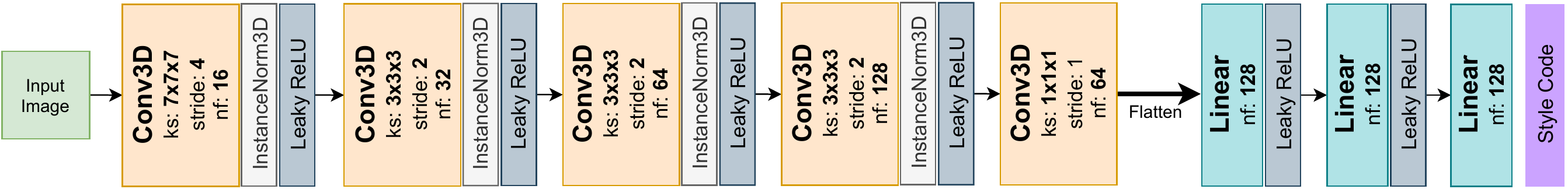}
    \caption{\textbf{The Architecture of the Style Encoder}.}
    \label{fig:style_encoder}
\end{figure}

Style Encoder consists of 3D convolutional layers along with 3D Instance Normalization \cite{ulyanov2016instance} and Parametric ReLU \cite{xu2015empirical} (Leaky ReLU) as the nonlinear activation with parameter $0.2$. The complete architecture is given in Fig.~\ref{fig:style_encoder}.

\subsubsection{Appearance Model's Architecture.} 
\begin{figure}[h]
    \centering
    \includegraphics[width=\linewidth]{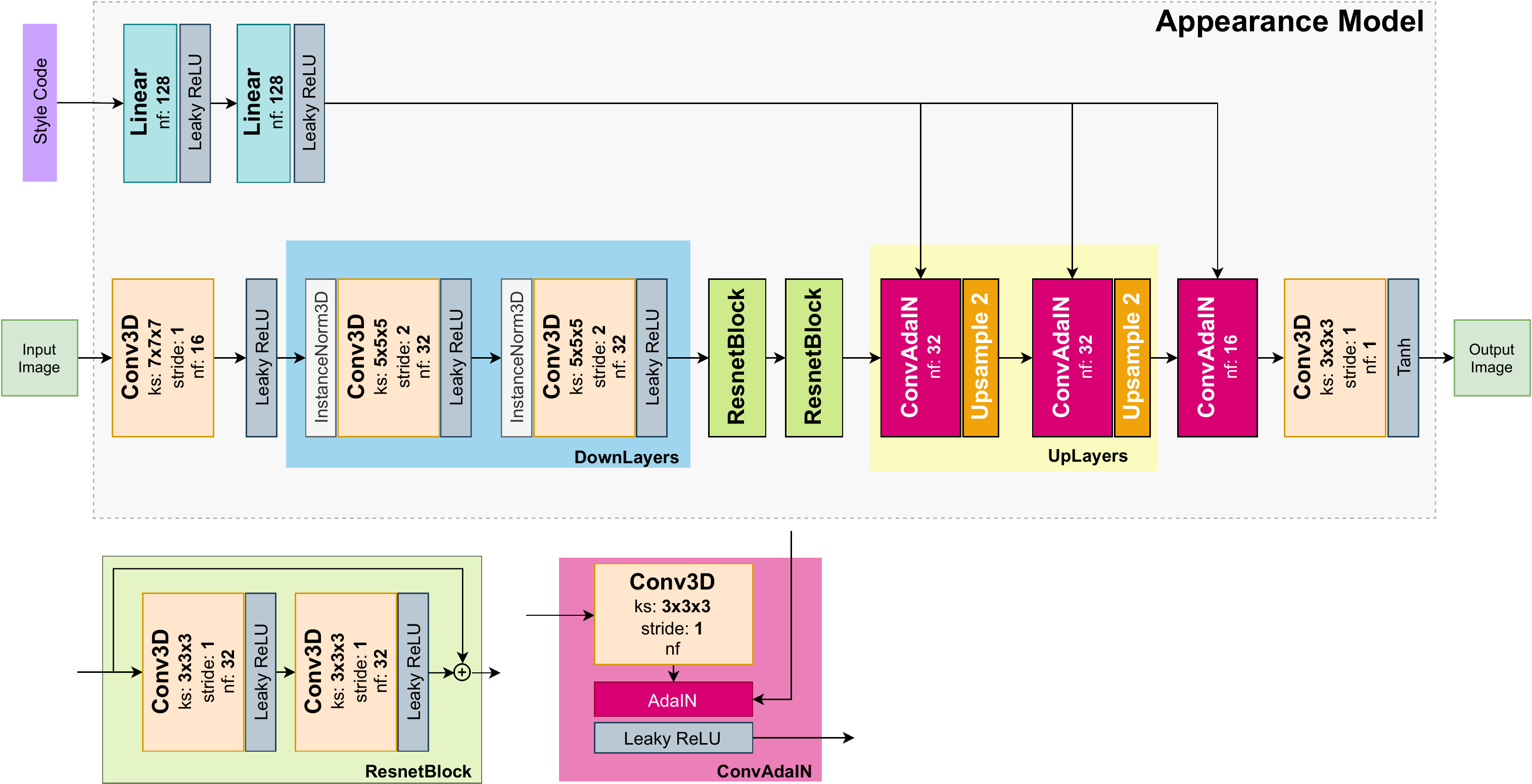}
    \caption{\textbf{The Architecture of the Appearance Model.} Appearance Model receives an \textbf{image} and \textbf{style code} and performs style manipulation of the input image based on the style code. \textbf{ResnetBlock} and \textbf{ConvAdaIn} blocks are shown separately.}
    \label{fig:appearance_model}
\end{figure}

The appearance model comprises 3D convolutions, Leaky ReLU activations with parameter $0.2$, and AdaIN \cite{huang2017arbitrary} layers. The architecture of the Appearance Model is given in Fig.~\ref{fig:appearance_model}.

\subsubsection{Flow Model's Architecture.} 
The Flow Model is a lighter version of 3D VoxelMorph \cite{balakrishnan2019voxelmorph} whose architecture is shown in Fig.~\ref{fig:flow_model}:

\begin{figure}[h]
    \centering
    \includegraphics[width=\linewidth]{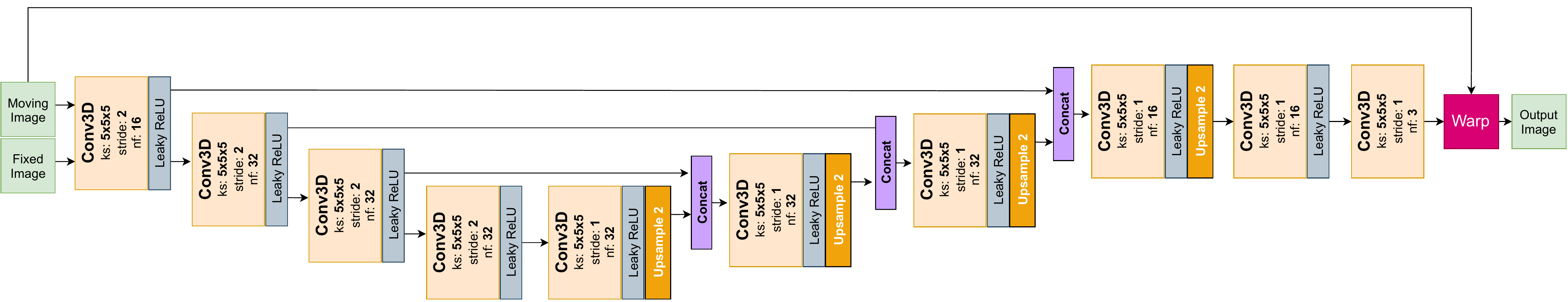}
    \caption{\textbf{The Architecture of the Flow Model.} We feed the Moving Image and Fixed Image to our Flow Model to obtain the corresponding flow field, which is then used to warp the Moving Image into the Fixed Image.}
    \label{fig:flow_model}
\end{figure}

\subsubsection{The Architecture of the Flow Adversarial Auto-Encoder}
\begin{figure}[h]
    \centering
    \includegraphics[width=\linewidth]{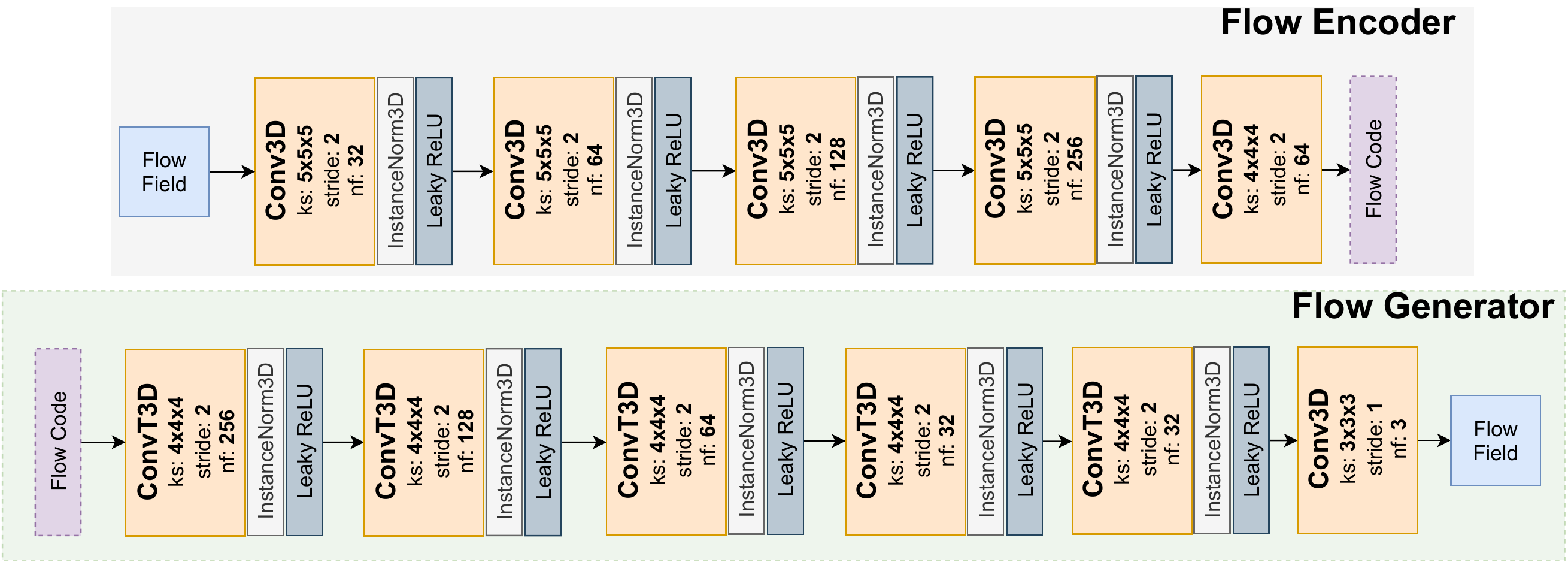}
    \caption{\textbf{The Architecture of the Flow AAE.} Flow Encoder encodes the given Flow Field into latent Flow Code while Flow Generator reconstructs the same Flow Field using the corresponding latent Flow Code. We regularize the Flow code using an adversarial loss.}
    \label{fig:aae_architecture}
\end{figure}

The architecture of the Flow Adversarial Autoencoder is shown in Fig.~\ref{fig:aae_architecture}.

\subsubsection{Latent Discriminators}
\begin{figure}[h]
    \centering
    \includegraphics[width=0.8\linewidth]{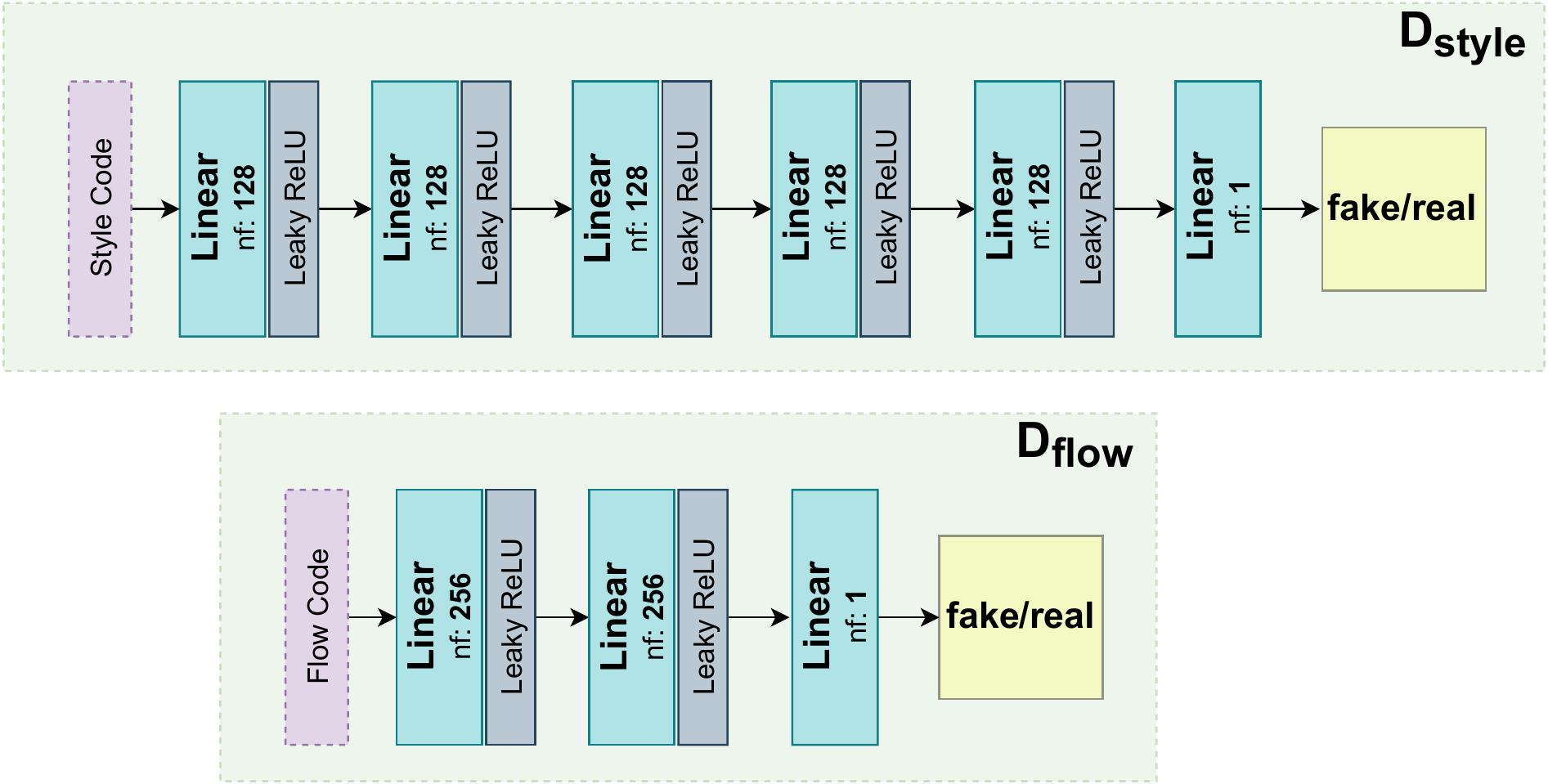}
    \caption{\textbf{The Architectures of the Latent Style and Latent Flow Discriminators.}}
    \label{fig:latent_discriminators}
\end{figure}

For the latent discriminators, we use fully connected multi-layer perceptrons. The architectures of Latent Style Code Discriminator $D_{style}$ and Latent Flow Code Discriminator $D_{flow}$ are shown in Fig.~\ref{fig:latent_discriminators}.

\subsubsection{3D U-Net}
\begin{figure}[h]
    \centering
    \includegraphics[width=\linewidth]{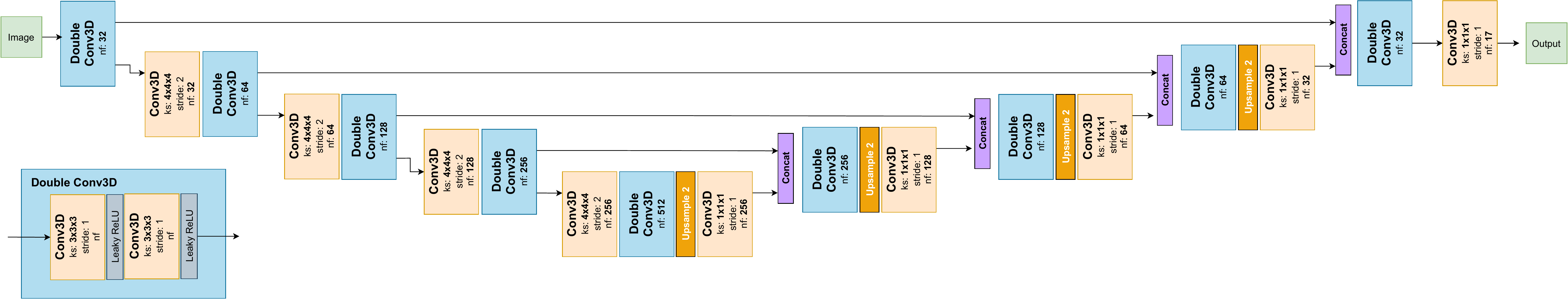}
    \caption{\textbf{The architecture of the 3D U-Net.}}
    \label{fig:unet_3d}
\end{figure}

The architecture of the 3D U-Net is shown in Fig.~\ref{fig:unet_3d}.

\subsection{Optimization Details for the Flow AAE}\label{AAE}
The Flow Adversarial Autoencoder (\textbf{Flow AAE}) corresponding to the base image is trained by optimizing the following objective:
\begin{equation*}
  \min_{\substack{E_{flow}\\G_{flow}}} \max_{D_{flow}} \mathbb{E}_{f\sim X_{flow}}\Big[\big\|f - G_{flow}(E_{flow}(f))\big\|_1+ \mu\\(D_{flow}(E_{flow}(f)) -1)^2\Big] + \mathbb{E}_{n\sim\mathcal{N}}\Big[\mu(D_{flow}(n))^2\Big]  
\end{equation*}
where $X_{flow}$ is the distribution of the flow field as generated by the Flow Model corresponding to the base image, $G_{flow}$ and $E_{flow}$ represents the Flow Decoder, and Flow Encoder of the Flow AAE model, $D_{flow}$ is the latent flow code discriminator, $\mathcal{N}$ is the normal distribution, and $\mu$ is the trade-off weight, and we set $\mu=0.1$.

\subsection{Additional Qualitative Results}\label{qualitative_results}

\subsubsection{Linear interpolation in the Flow Latent Space} 
\begin{figure}[h]
\begin{center}
\includegraphics[width=0.9\linewidth]{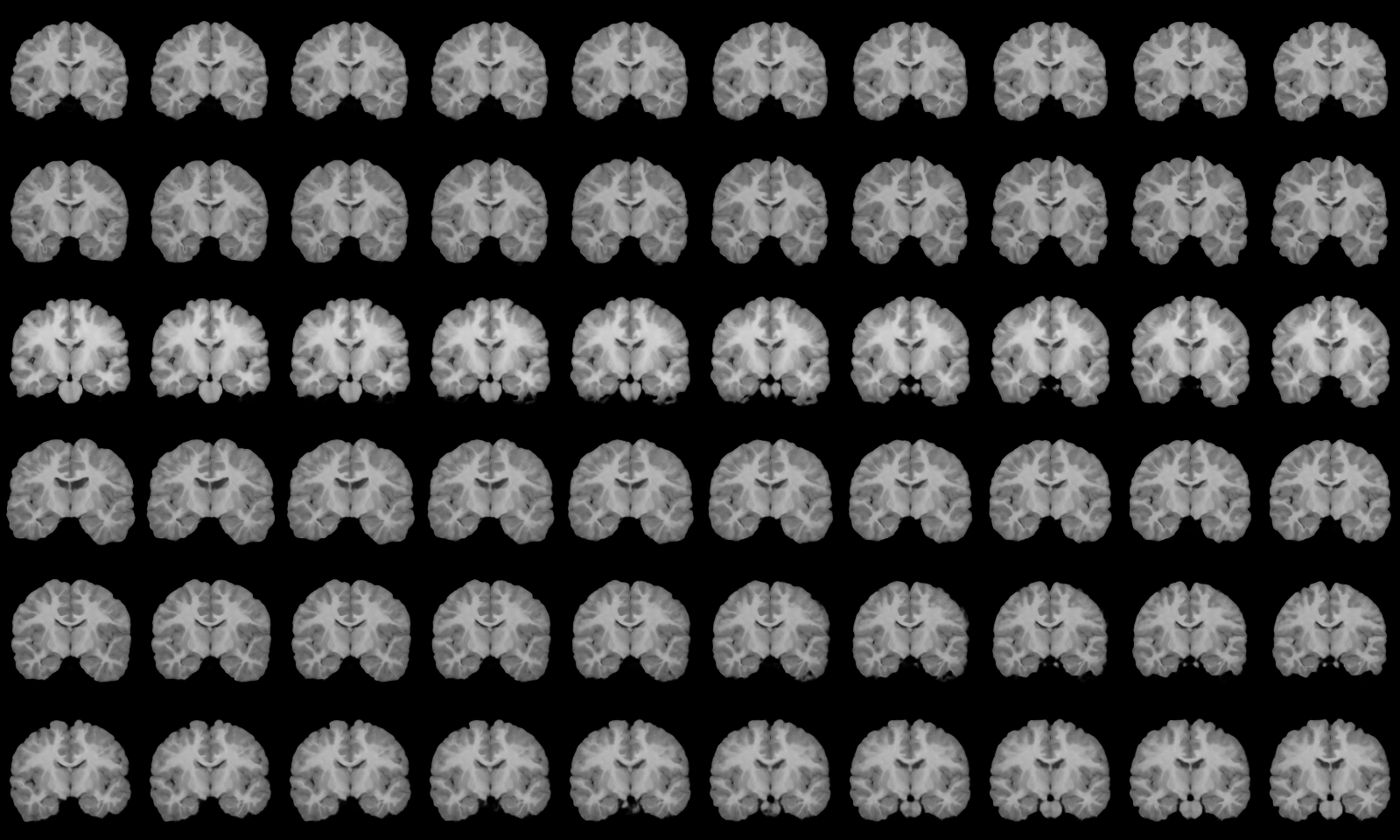}
\includegraphics[width=0.9\linewidth]{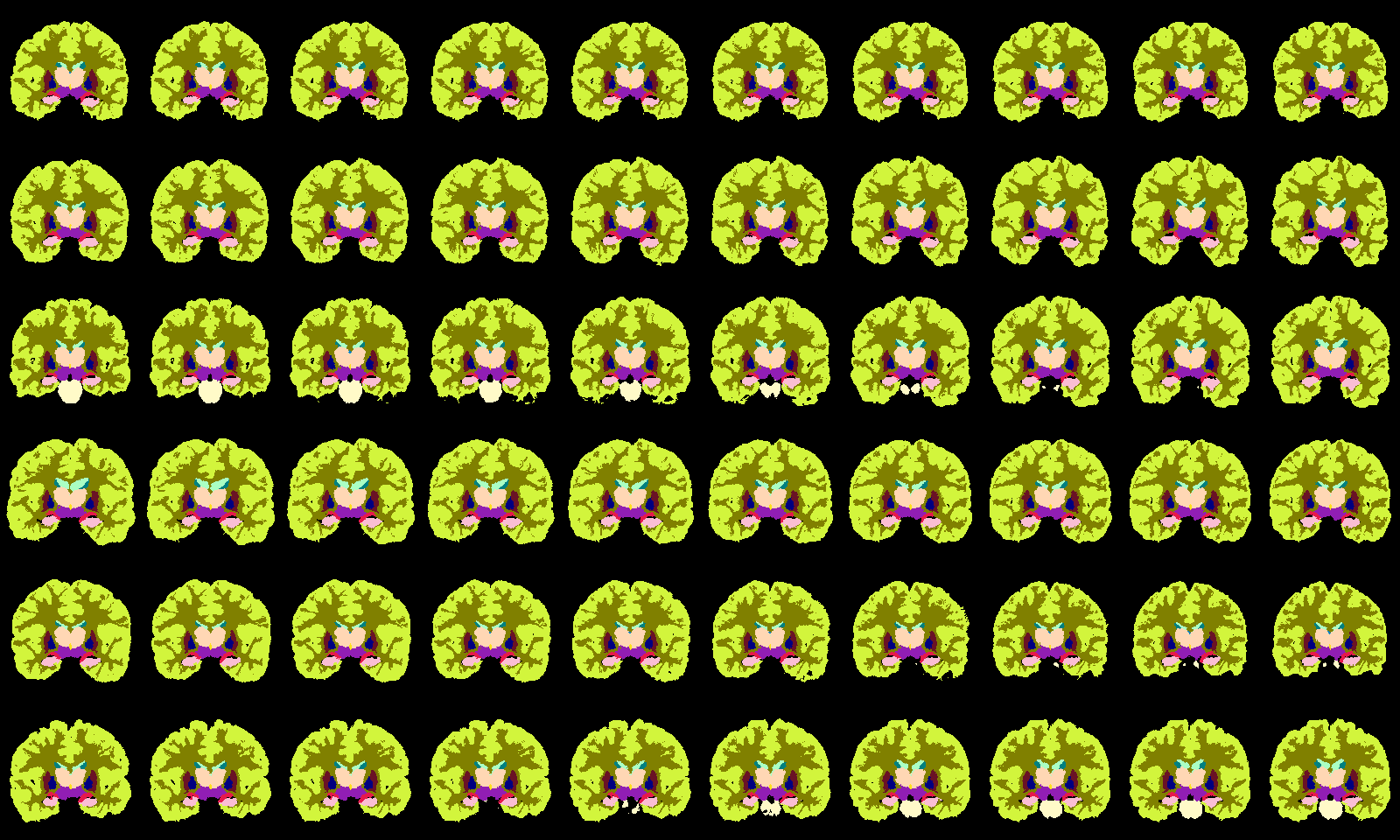}
\end{center}

\caption{\textbf{A linear walk in the Flow Latent Space.} Images and segmentation labels are generated by linearly interpolating between two different flow latent codes. \textit{Left to right:} Linear interpolation of the flow latent codes from the first column to the last column using the same style.}
\label{fig:linear_walk}
\end{figure}

Fig.~\ref{fig:linear_walk} shows the efficacy of our proposed method in obtaining a linear Flow Latent Space. We observe a smooth transition of the image-segmentation pairs generated using a convex combination of two different latent flow codes.

\subsubsection{The t-SNE Projection of the Style Codes}
\begin{figure}[h]
    \centering
    \includegraphics[width=\linewidth]{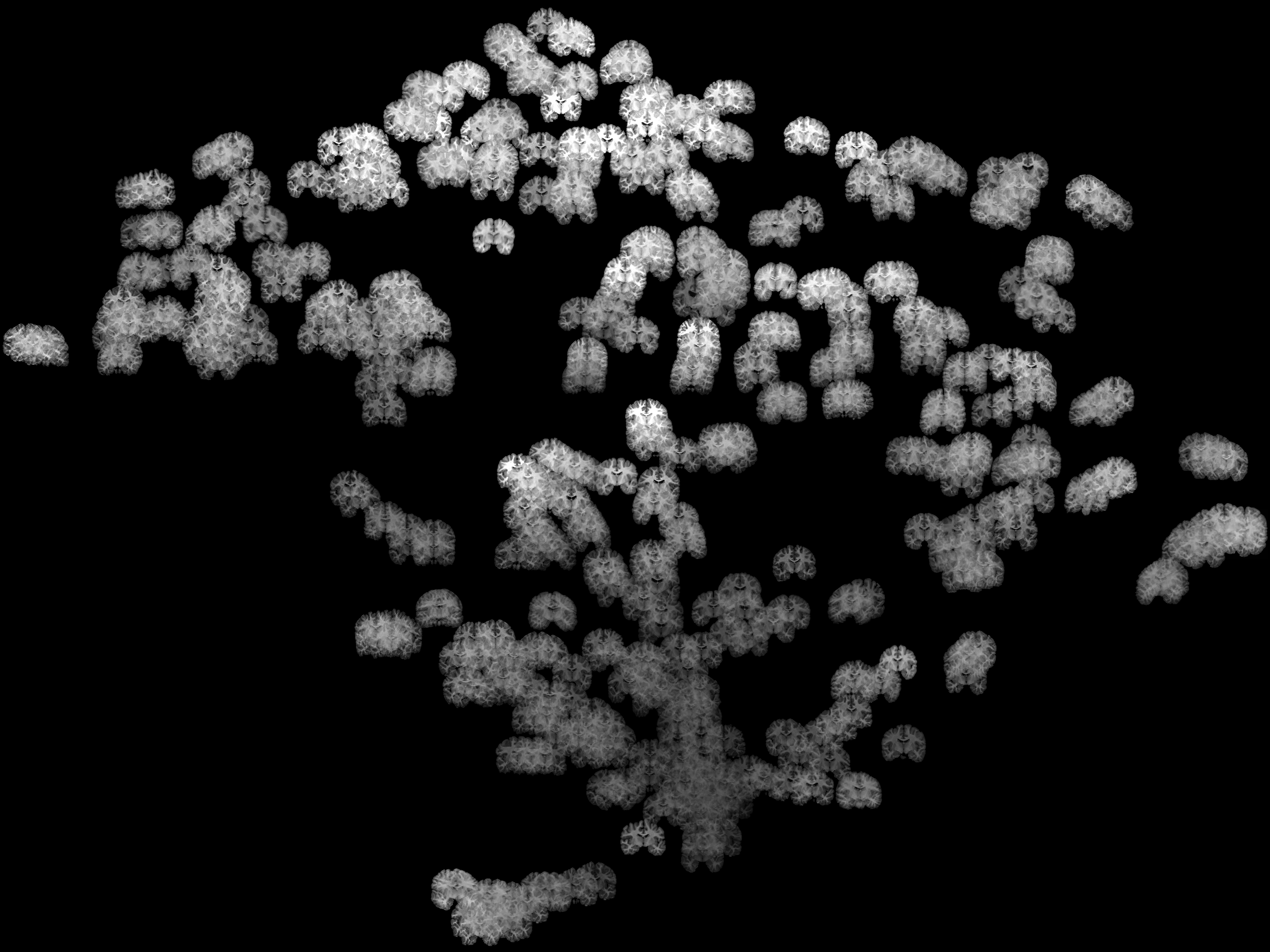}
    \caption{\textbf{The t-SNE 2D projection of the style codes.} Self-supervised clustering of similar styled images is shown.}
    \label{fig:tsne_projection}
\end{figure}

Fig.~\ref{fig:tsne_projection} shows a 2D projection of the style latent codes and corresponding images obtained by our method using self-supervised volumetric contrastive loss. We observe that images with a similar style are clustered together.

\subsubsection{Flow Fields Examples}
\begin{figure}[t]
\begin{center}
\begin{tabular}{@{}ccc@{}}
     \includegraphics[width=0.15\linewidth]{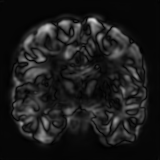}
    \includegraphics[width=0.15\linewidth]{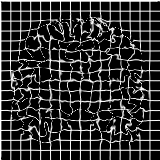} & \includegraphics[width=0.15\linewidth]{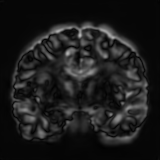}
    \includegraphics[width=0.15\linewidth]{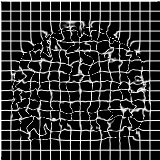} & \includegraphics[width=0.15\linewidth]{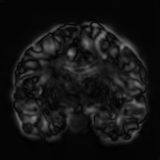}
\includegraphics[width=0.15\linewidth]{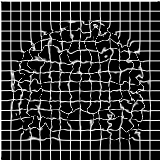}\\
     (a) & (b) & (c)
\end{tabular}
\end{center}
   \caption{\textbf{Deformation fields applied on grid images.} \textit{Left}: norm of vector flow fields. \textit{Right}: flow field applied on grid image.}
\label{fig:l1loss}
\end{figure}

Fig.~\ref{fig:l1loss} shows sample results of the deformation field applied on grid images.

\end{document}